\documentclass[a4j,10pt]{article}
\usepackage{amsmath,amsthm,amssymb}
\usepackage{enumerate}
\title{Generation of Two-Layer Monotonic Functions}
\author{Yukihiro KAMADA and Kiyonori MIYASAKI}
\date{}
\begin{document}
\maketitle

\begin{abstract}
The problem of implementing a class of functions with particular conditions by using monotonic multilayer functions is considered.
A genetic algorithm is used to create monotonic functions of a certain class, and these are implemented with two-layer monotonic functions.
The existence of a solution to the given problem suggests that from two monotone functions, a monotonic function with the same dimensions can be created.
A new algorithm based on the genetic algorithm is proposed, which easily implemented two-layer monotonic functions of a specific class for up to six variables.
\end{abstract}


\section{Introduction}
For any given Boolean function to be a threshold function, it must satisfy monotonicity.
A monotonic multilayer network using monotonic functions as elements of the network should be more capable of processing information than the threshold-element networks that have been used till now because the set of all monotonic functions includes the set of all threshold functions.
To construct monotonic multilayer networks, besides using monotonic functions, a method of combining AND, OR, and NOT functions has also been proposed\cite{Muroga}.
Although this method makes network construction easy, replacing one element requires the reconstruction of the entire network.
Therefore, in this study, to implement various logical functions using preexisting networks, a monotonic multilayer network using EXOR function is proposed.

The monotonic multilayer functions that enable the operation of monotonic multilayer networks can implement an arbitrary logical function, however, the characteristics of the function are not well known.
First, we consider the problem of implementing a class of functions with particular conditions by using monotonic multilayer functions.
In particular, a genetic algorithm is used to create monotonic functions of a certain class, and these are implemented with two-layer monotonic functions.
The existence of a solution to the given problem suggests that from two monotone functions, a monotonic function with the same dimensions can be created.
In addition, we propose a new algorithm based on the genetic algorithm, which easily implemented two-layer monotonic functions of a specific class for up to six variables.

\section{Preliminaries}
The following definition is due to Ref. 2.
\subsection{Monotonic Functions}
A boolean function in the variables $x_1$, $x_2, \ldots $, $x_n$ is a map $f : \left\{ 0,1 \right\}^n \rightarrow \left\{ 0,1\right\}$.
The collection of all $n$-ary boolean functions is denoted $\mathcal{B}^n$.
\textit{Monotonic functions} are defined as the class of functions $f \in \mathcal{B}^n$ satisfying \[x \preceq  y \Rightarrow f(x) \leq f(y), \]
where for $n$ tuples $ x = ( x_1, x_2, \ldots , x_n ) $, $ y = ( y_1, y_2, \ldots , y_n ) $ we define $x \preceq y$ if and only if $x_i \leq y_i$, $\forall i \in \left\{ 1, 2, \ldots , n \right\}$.
\subsection{Monotonic Multilayer Functions}
The set of monotonic functions is expressed as $\mathcal{M}$.
\textit{Monotonic multilayer functions} are defined as the class of functions that satisfy the following equation:
\[
f(x) = f_1(x) \oplus f_2(x) \oplus \cdots \oplus f_k(x),
\]
where $f \in \mathcal{B}^n$ and $\left\{f_1, f_2, \ldots , f_k \right\} \in \mathcal{M}$ for $x \in \left\{0, 1 \right\}^n$ .
\subsection{Problem}
We consider the problem of forming two-layer monotonic functions as an example of analyzing the structure of monotonic multilayer networks.
In particular, a genetic algorithm will be used to form two $n$-variable monotonic functions for obtaining a monotonic function with the same number of dimensions as these two functions.
\subsection{Expression of Solution and Genetic Algorithm}
The solution to the problem mentioned above will be given by gene g, as follows:
\[
g=
\underbrace{f(X_1), f(X_2), \ldots , f(X_{2^n}), h(X_1), h(X_2), \ldots , h(X_{2^n})}_{2\times 2^n bits}
\]
where $X_j \in \left\{0, 1 \right\}^n $, $j \in \left\{1, 2, \ldots , 2^n \right\}$.

The processes of the genetic algorithm are laid out as follows:
There are 1000 genes prepared that are initialized as 0(1).
Tournament selection is used for selecting the genes.
To prevent a collapse into localized solutions as much as possible, the selected number of genes is set to 2.
Unified crossovers are used.
Mutation probability is set to 0.01.
\begin{table}[htbp]
\begin{center}
\begin{tabular}{c|c|c} \hline
Functions & $true$ & $false$ \\ \hline
$f$ & 10 & 1 \\ 
$g$ & 10 & 1 \\ 
$f \oplus g$ & 10 & 1 \\ \hline
$c(f)$ & 1 & 10 \\ 
$c(g)$ & 1 & 10 \\ 
$c(f \oplus g)$ & 1 & 10 \\ \hline
\end{tabular}
\end{center}
\caption{Parameters Necessary for Calculating Fitness}
\end{table}

Table 1 lists the parameters needed to calculated the appropriate genes (their fitness).
Rows 2-4 of the table indicate that if $f$, $g$, or $f \oplus g$ represent each monotonic function, 10 points will be added to their fitness value; else, 1 point will be added to their fitness value.
Rows 5-7 show that if $f$, $g$, or $f \oplus g$ represent each constant function one point will be added to their fitness value; else, 10 points will be added. This evaluation prevents function degeneracy.
Thus, the maximum value of fitness is 60 points.

\section{Calculation Results}
\begin{table}[htbp]
\begin{center}
\begin{tabular}{c|c|c} \hline
Number of Variables & Generation Numbers & Fitness \\ \hline
2 & 50 & 60 \\ 
3 & 50 & 60 \\ 
4 & 50 & 60 \\ 
5 & 1000 & 33 \\ 
6 & 10000 & 33 \\ \hline
\end{tabular}
\end{center}
\caption{Calculation Results Using a Genetic Algorithm}
\end{table}
The Column 1 of Table 2 shows the number of variables for the two-layer monotonic function that is produced.
Column 2 shows the maximum value for generation numbers in the genetic algorithm.
Column 3 shows the maximum value of fitness gained for each generation number indicated in Column 2.
For up to four variables, it was possible to obtain monotonic functions that could implement two-layer monotonic functions.
In the case of five or six variables, the monotonic functions that will act as network elements could not be obtained, and a monotonic function with the same dimensions could not be produced.
This is possibly to due to a very large search space.

\section{Improved Algorithm}
By using the genetic algorithm, we attempted to generate a same-dimension monotonic function by using the functions of the same class, but as the variables increased, we could not to achieve this goal.
Thus a different algorithm, as described below, was proposed.
For expressing the solution to this problem, as in the case of the genetic algorithm, we used genes. However, in this case we used four genes.

\begin{enumerate}[Step 1.]
     \item Initialize genes at 0, and express it as $g_0$.
     \item Calculate the fitness value of $g_0$, and express it as $e_0$.
     \item The genes among $g_0$ that reverse only the arbitrary three bits as 0¨1 (1¨0) are $g_1$. The fitness value is $e_1$.
     \item The genes among $g_0$ that reverse only the arbitrary two bits and 1 bit are $g_2$ and $g_3$, respectively. Their fitness values are $e_2$ and $e_3$, respectively.
     \item If out of $e_1$, $e_2$, and $e_3$, the maximum fitness exceeds $e_0$, the corresponding gene becomes a new gene, $g_0$.
\end{enumerate}

From this stage onwards, repeat Steps 2-5 for the predetermined number of times.
The parameters and calculation method of fitness are the same as those used with the genetic algorithm.

\section{Calculation Results Using Our Algorithm}
\begin{table}[htbp]
\begin{center}
\begin{tabular}{p{5em}|p{8em}} \hline
Number of Variables & Number of times  Steps 2-5  \\ \hline
\hfil 2 & \hfil 2  \\ 
\hfil 3 & \hfil 1  \\ 
\hfil 4 & \hfil 4  \\ 
\hfil 5 & \hfil 47  \\ 
\hfil 6 & \hfil 185  \\ \hline
\end{tabular}
\end{center}
\caption{Calculation Results Using Our Algorithm}
\end{table}
The Column 1 of Table 3 is the same as the Column 1 of Table 2.
Column 2 indicates the number of times Steps 2-5 of our algorithm was repeated until the maximum fitness value was attained.
For the given problem, an optimum solution was obtained for cases with up to six variables.

\section{Considerations}
Similar to elitism, our algorithm updates the genes only when their fitness values exceed current conditions and thus genes are not degraded because of decreasing fitness values. 
Because, gene reversal occurs for three bits at the most, local solutions are more easily avoided for each calculation.
In the case of a genetic algorithm, the generation numbers and gene numbers must be adjusted in certain cases.
For the adjustments in this study, the results would easily fall into local solutions.
Our new algorithm apparently to makes it slightly easier to reach the optimum solution.
In practice, in multiple computer experiments, optimum solutions were reached for all attempts.

\section{Conclusions}
A genetic algorithm was used to form the two-layer monotone functions of a certain class.
For up to four variables, a solution was easily obtained.
In the case of five or six variables, a new algorithm was proposed to effectively construct the network.
We showed that a function of the same class could be obtained from two monotone functions by implementing monotonic functions with a monotone multilayer network.
The formation of the functions could become a foothold in performing the structural analyses of networks.
As our future initiatives, we will inspect systematic methods to generate multilayer functions and adapt our algorithm to various problems to perform a comparative review.

\end{document}